\documentclass[letterpaper, 10 pt, journal, twoside]{IEEEtran}
\usepackage{amsmath,amsfonts}
\usepackage{algorithm}
\usepackage{bm}
\usepackage[noend]{algpseudocode}  
\usepackage{algorithmicx}

\usepackage{multirow}
\usepackage{booktabs}
\usepackage{array}
\usepackage[caption=false,font=normalsize,labelfont=sf,textfont=sf]{subfig}
\usepackage{textcomp}
\usepackage{stfloats}
\usepackage{url}
\usepackage{verbatim}
\usepackage{graphicx}
\usepackage{cite}
\begin{document}

\markboth{IEEE Robotics and Automation Letters. Preprint Version. Accepted March, 2024.}
{Lin \MakeLowercase{\textit{et al.}}: Modular Multi-Level Replanning TAMP Framework for Dynamic Environment}

\author{Tao Lin, Chengfei Yue, Ziran Liu, and Xibin Cao%
\thanks{Manuscript received: October, 11, 2023; Revised January, 31, 2024; Accepted March, 7, 2023.}

\thanks{This letter was recommended for publication by Editor Tamim Asfour upon evaluation of the Associate Editor and Reviewers' comments.
This work was supported by National Natural Science Foundation of China 12372045 under Shenzhen Science and Technology Program JCYJ20220818102207015. (Corresponding author: Chengfei Yue.)}

\thanks{Tao Lin, Ziran Liu, and Xibin Cao are with Research Center of the Satellite Technology, Harbin Institute of Technology, Harbin 150001, China (e-mail: lintao6324@gmail.com; lzr1102@gmail.com; xbcao@hit.edu.cn)}%

\thanks{Chengfei Yue is with Institute of Space Science and Applied Technology, Harbin Institute of Technology Shenzhen, Shenzhen 518055, China (e-mail: yuechengfei@hit.edu.cn)
}
\thanks{Digital Object Identifier (DOI): see top of this page.}
}

\title{Modular Multi-Level Replanning TAMP Framework for Dynamic Environment}

\maketitle

\begin{abstract}
Task and Motion Planning (TAMP) algorithms can generate plans that combine logic and motion aspects for robots. However, these plans are sensitive to interference and control errors. To make TAMP algorithms more applicable and robust in the real world, we propose the \underline{m}odular \underline{m}ulti-level \underline{r}eplanning TAMP \underline{f}ramework(MMRF), expanded existing TAMP algorithms by combining real-time replanning components. MMRF generates an nominal plan from the initial state and then reconstructs this nominal plan in real-time to reorder manipulations. Following the logic-level adjustment, MMRF attempts to replan a new motion path, ensuring that the updated plan is feasible at the motion level. Finally, we conducted several real-world experiments. The result demonstrated MMRF swiftly completing tasks in scenarios with moveing obstacles and varying degrees of interference. 
\end{abstract}


\section{Introduction}
\IEEEPARstart{I}{n} order to solve complex tasks autonomously, robots need to plan discrete subtask plans at the logic level, and also find corresponding continuous motion paths for each subtask to ensure their feasibility at the motion level. Task and motion planning (TAMP)\cite{srivastava2014combined,garrett2021integrated} algorithms combining high-level task planning and bottom-level motion planning can effectively solve this class of problems. Among them, sampling-based TAMP algorithms\cite{srivastava2014combined,garrett2018ffrob,garrett2018sampling,garrett2020online,garrett2020pddlstream} have been widely studied because of their probabilistic completeness.

\begin{figure}[htbp]
\centering
\subfloat[]{\includegraphics[width=0.3\columnwidth]{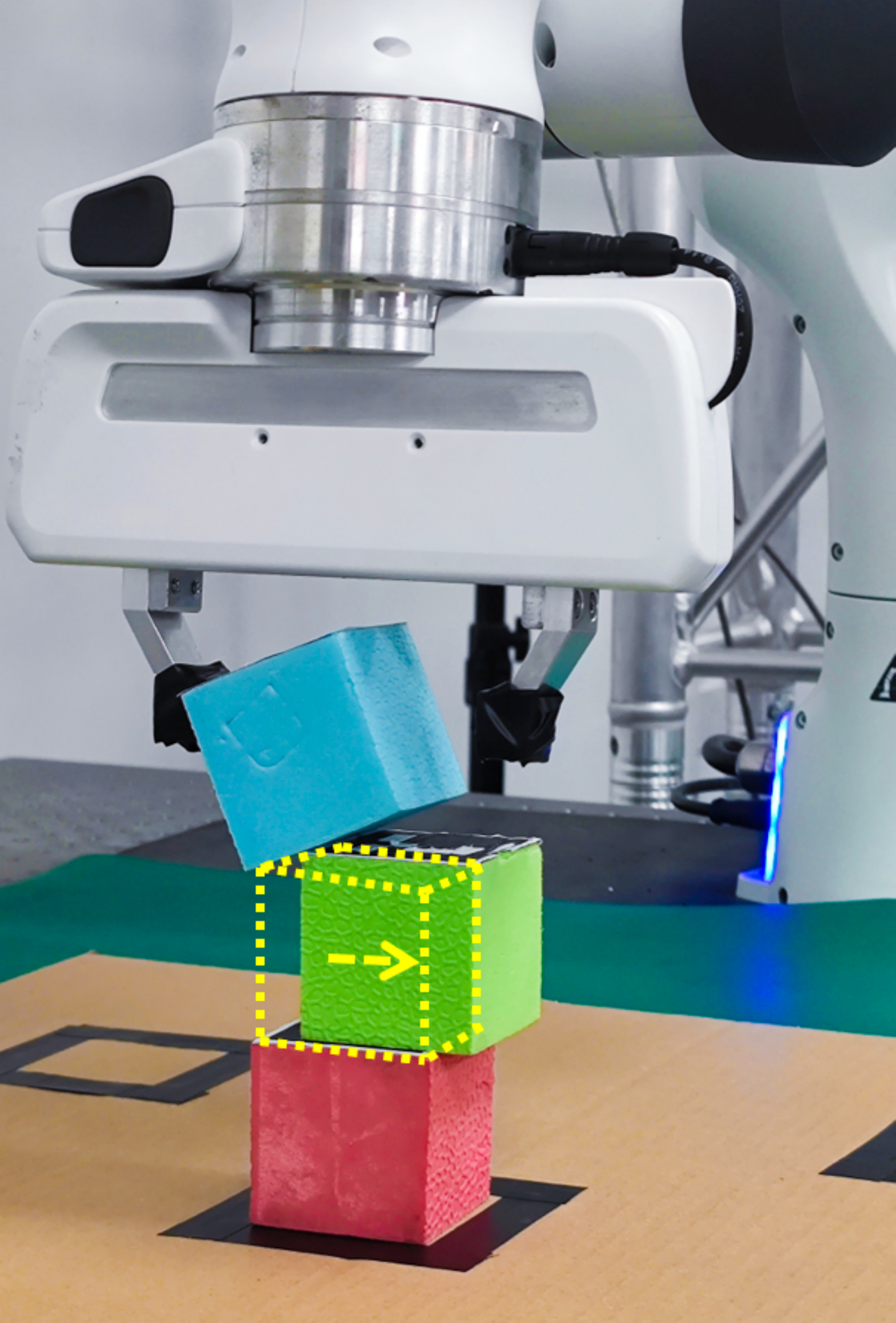}
\label{slight}}
\hfil
\subfloat[]{\includegraphics[width=0.3\columnwidth]{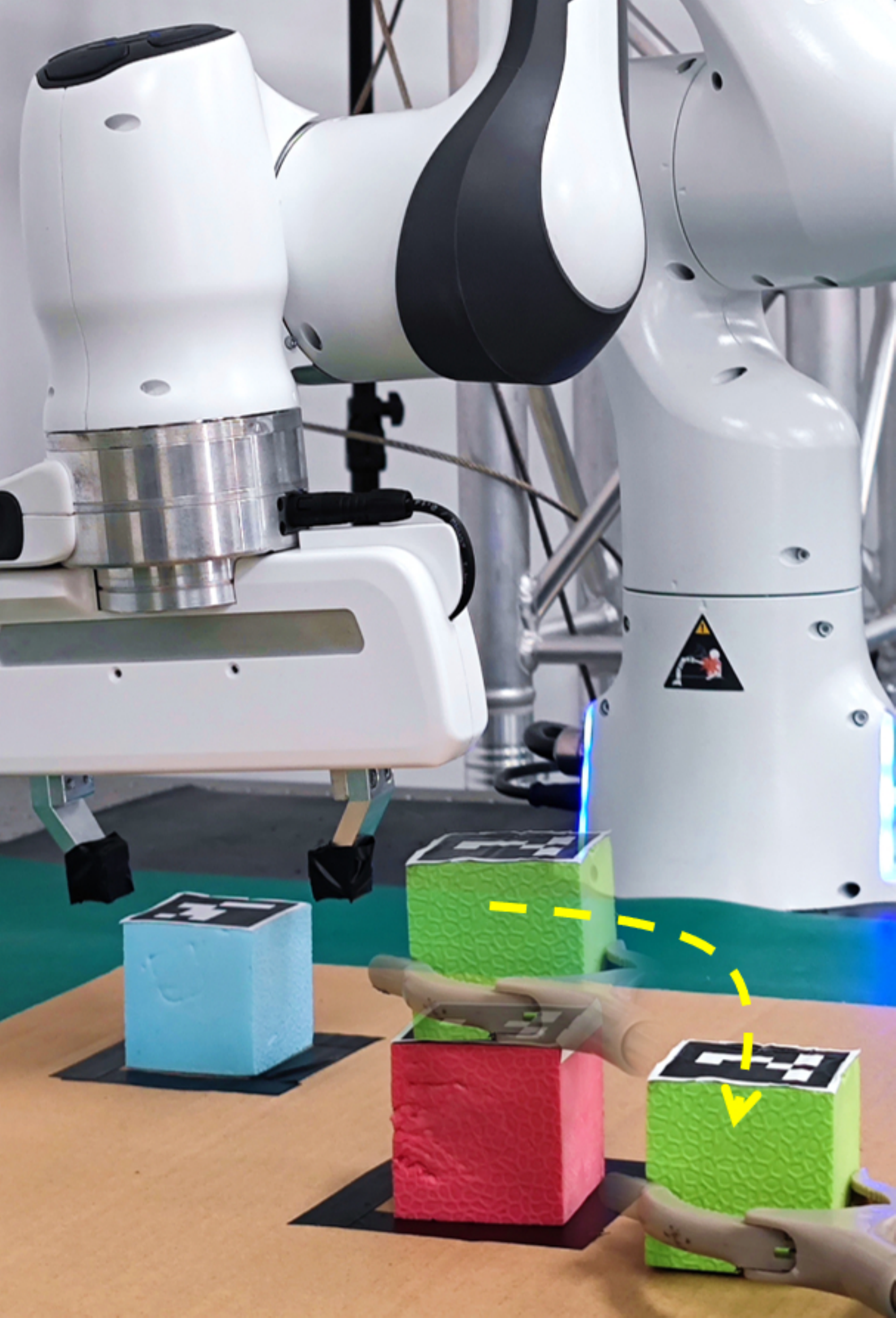}
\label{middle}}
\hfil
\subfloat[]{\includegraphics[width=0.3\columnwidth]{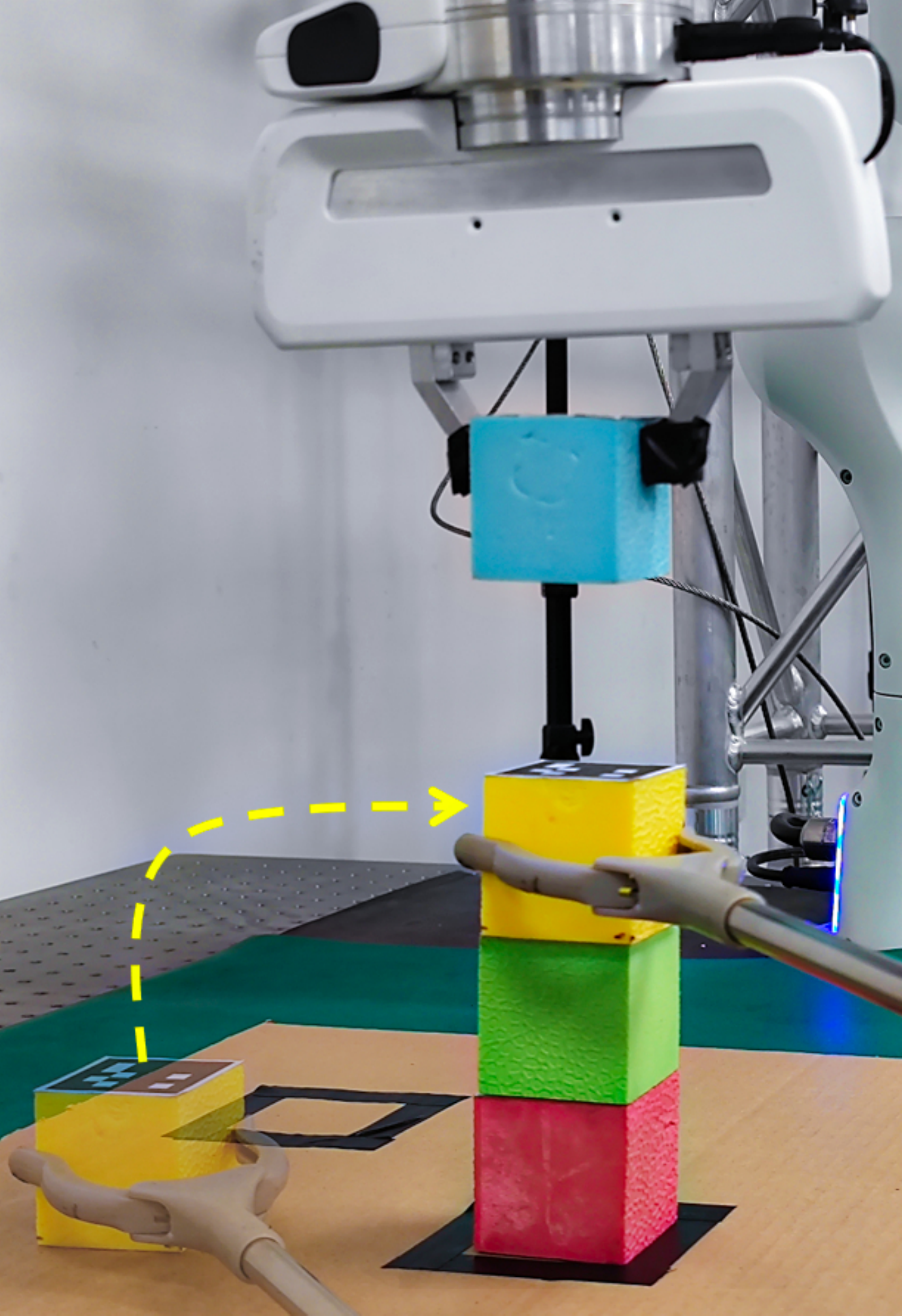}
\label{heavy}}

\caption{Three levels of interference. (a) Slight interference does not affect the execution order of the plan, but requires motion replan for placing. (b) Middle interference disrupts the sequence of plan. Robot needs to re-pick and re-place the green block. (c) Heavy interference make the plan invalid ,and can only be resolved by TAMP replanning}
\label{interference}
\end{figure}

However, the plans generated by sampling-based TAMP algorithms are only valid in static environment and under highly accurate control. Any errors may make plans unexecutable and, in serious cases,  may even damage the robot. For the sake of further discussion, we collectively refer to environmental variations and control errors as \textbf{interference}, which is categorized as \textbf{slight}, \textbf{middle} and \textbf{heavy} according to its effects on the plans. Taking the stack task domain as an example, the robot needs to stack the red, green and blue blocks in sequence.

\textbf{Slight interference}, which does not affect the execution order of the original plan, but requires motion replanning. This interference is typically a slight displacement of the objects, which is usually caused by control errors and is prevalent in contact-rich operations. In Fig.\ref{interference}\subref{slight}, the green block was not placed in the desire position. This interfere make robot fail to move along the original motion path to place the blue block.

\textbf{Middle interference}, disrupts the sequence of the original plan, but the plan is still feasible. The robot needs to adjust the execution order of the plan and then replan the motion path. In Fig.\ref{interference}\subref{middle}, when the robot is about to pick the blue block, the green block is moved down. At this point, the robot needs to backtrack to the previous subtask to re-pick and re-place the green block.

\textbf{Heavy interference}, will make the original plan completely invalid. The robot needs to perform TAMP full planning again. In Fig.\ref{interference}\subref{heavy}, when the robot is about to place the blue block, the yellow block is placed to obstruct its path. Since the original plan only includes the picking and placing of the blue and green blocks, it does not include the operations of the other blocks. So robot needs perform TAMP replanning to generate a new plan for the current state.

Theoretically, TAMP replanning can solve the all above interference. However, TAMP planning is time-consuming. If the environment changes during the planning process, the new plan may be invalidated again, leading to endless replanning. Therefore, during execution, the robot needs to have reactive planning capabilities to cope with the first two levels of interference. Additionally, some work extends the TAMP algorithm so that it can be used in practice, but it can only cope with a fraction of the interference.

We propose a modular multi-level replanning TMAP framework(MMRF) that incorporates the TAMP Solver, Subtask Scheduler and Subtask Planner, Robot Controller, and State Evaluator. After the TAMP Solver generates a nominal subtask plan, the Subtask Scheduler rebuilds the nominal subtask plan online based on the logic state. Then, the Subtask Planner is notified to perform motion planning for the subtasks in the actual plan to verify their feasibility at the motion level.
Finally, the Robot Controller performs higher frequency reactive control based on the results generated by the Subtask Planner. The TAMP Solver will only replan when the Subtask Scheduler or Subtask Planner fails, minimizing the number of time-consuming TAMP planning and improving the rapid response ability to interference.

Our contributions are:

1) We present a Modular Multi-level Replanning TAMP Framework(MMRF). MMRF can perform real-time low-level replanning based on existing plans at the logic and motion. This framework enables robots to quickly respond to interference and reduce the number of time-consuming TAMP planning.

2) MMRF provides a stable paradigm for the application of current TAMP algorithms in the real world. The decoupled modular design makes the framework can combine different TAMP algorithms and motion planning algorithms.

3)We apply MMRF to the real-world stack and rearrange task domains and validate its effectiveness on the Franka Emika Panda\footnote{\url{https://sites.google.com/view/mmrf/project}}. The results show that MMRF can accomplish tasks quickly under moving objects and various interference with a shorter completion time compared to previous frameworks.

\section{RELATED WORK}

\subsection{Task and Motion Planning}
The concept of Task and Motion Planning (TAMP) has evolved and enriched over the years of research. Sampling-based algorithms\cite{srivastava2014combined,garrett2018ffrob,garrett2018sampling,garrett2020online,garrett2020pddlstream}, also known as TAMP algorithms in a narrow sense, are mostly based on the Planning Domain Definition Language (PDDL)\cite{fox2003pddl2} that defines state and skill primitives as discrete abstract symbols. The algorithms sample discrete symbols at the logic level and find feasible plans by traditional symbolic planners such as FastDownward\cite{helmert2006fast}, and then sample feasible motion paths in a continuous space for validation. The algorithms iterative sample between both levels until they find a plan that is both logic and motion feasible. They have the advantage of probabilistic completeness, but their plans rely on static environments and precise controls. There is also a lot of work extending sample-based algorithms. In \cite{yang2022sequence,kim2022representation}, neural networks were introduced to accelerate planning. Learning skills can help the TAMP system adapt to various scenarios \cite{wang2021learning,liu2023synergistic}. In \cite{curtis2022long}, the integration of neural networks in state evaluation enhances the system to operate unknown objects and complete long-term tasks.

There are also optimization-based methods, among which logic-geometric programming(LGP)\cite{toussaint2015logic,toussaint2018differentiable,toussaint2020describing} is the most representative. In LGP, subtasks are transformed into geometric constraints. The entire motion path is optimally planned under these discrete constraints, while sampling-based TAMP methods independently sample motion paths for each subtask. When LGP is used for TAMP, it requires manually specifying the subtask sequence or combining it with other task planning method. Therefore, LGP can serve as the Subtask Planner in our framework. After planning the actual plan, the Subtask Scheduler converts it into discrete geometric constraints for LGP to perform motion planning.

With the development of large language models(LLM), several LLM-based frameworks\cite{ahn2022can,lin2023text2motion,driess2023palm,brohan2023rt} have emerged in recent years. Among these, RT-2\cite{brohan2023rt} demonstrates significant generality across various scenes and tasks. RT-2 is an end-to-end architecture that takes input in the form of images and natural language, directly outputting end-effector actions. However, these LLM-based methods require significant computational power and are generally not suitable for local computations. The planning frequency is relatively low, usually between 1-5 Hz, leading to noticeable end-effector jitters.

Our framework aims to extend sampling-based TAMP algorithms, enabling rapid replanning of the original plan based on the newest state. This allows for quick responses to interference, minimizing the need for time-consuming TAMP replanning. The modular and decoupled design also provides the framework with the potential to be compatible with the methods mentioned above.

\subsection{Online replanning}
Reactive control can effectively address slight interference. Migimatsu and Bohg\cite{migimatsu2020object} propose object-centered reactive control, which allows the original plan to remain valid after the objects have been displaced. However, reactive control can only resist interference at the motion level and will fail if the sequence of nominal plan is disrupted.

There is also much work on maintaining the execution flow of the plan at the logic level, such as state machines\cite{bohren2010smach} and behavior trees\cite{hannaford2018behavior}, but they are complex and difficult to define and expand. Paxton et al.\cite{paxton2019representing} proposes a robust logical dynamic system (RLDS) that can flexibly transform the nominal plan into a chain of logical dynamics. The order of task execution is quickly adjusted according to the logic state , which can effectively solve slight and middle interference. This serves as inspiration for the Subtask Scheduler in our framework. However, these algorithms only maintain the plan at the logic level and do not verify the motion feasibility of the whole plan. 

Garrett et al.\cite{garrett2020online} propose an online replanning approach to cope with real-world uncertainty. After executing a non-precise action or after the state of an object has been updated, the framework quickly performs TAMP replanning using the original plan. This approach is able to effectively cope with all three kinds of interference, but even slight interference can cause time-consuming replanning. Our framework adopts the idea of using the original plan to accelerate TAMP replanning.

\section{Modular Multi-Level Replanning Framework}

\begin{figure*}[!t]
\centering
\includegraphics[width=2.0\columnwidth]{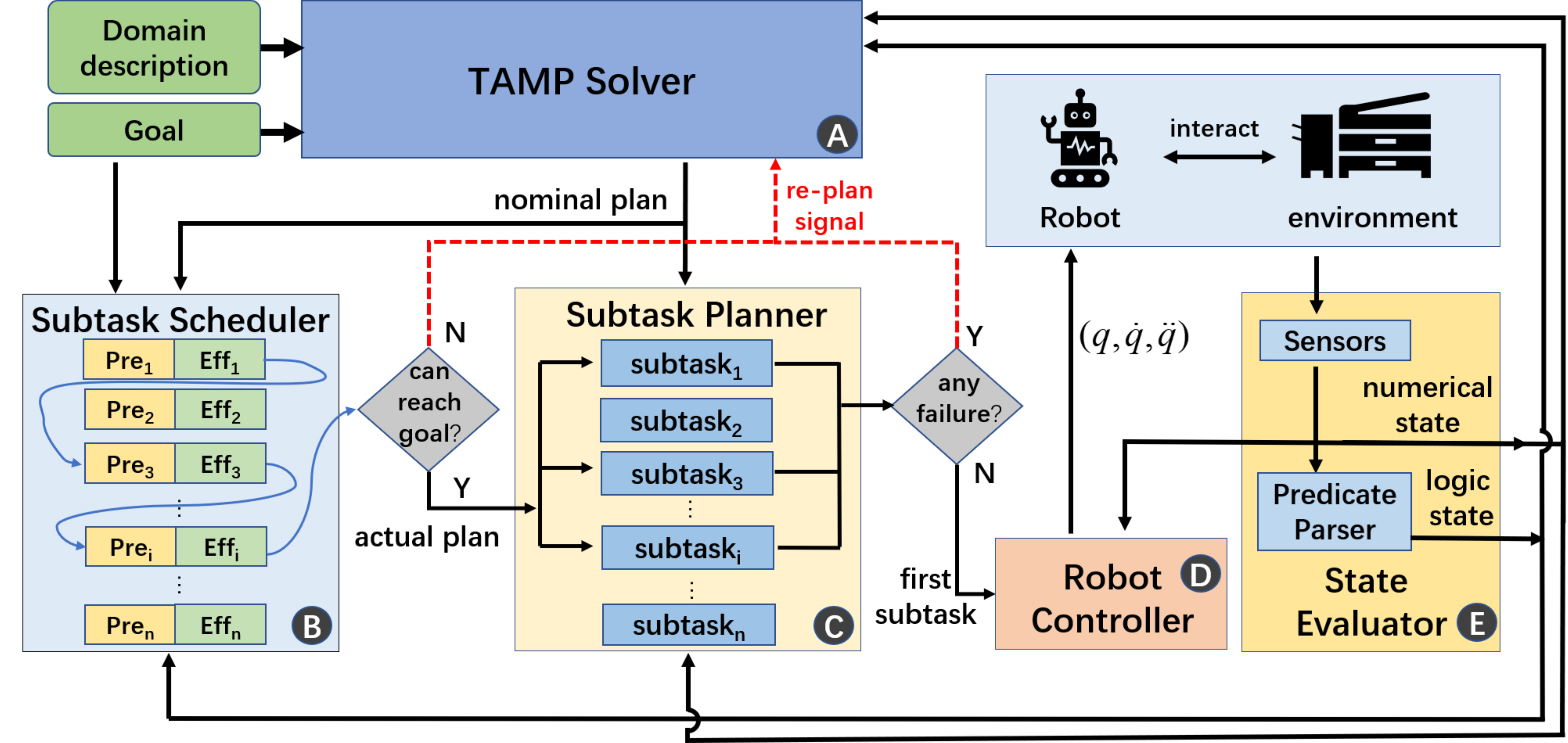}
\caption{Structure of the modular multi-Level replanning TAMP framework. (a) TAMP Solver:generates nominal plans. (b) Subtask Scheduler:Reconstructs the nominal plan into the actual plan to ensure the feasibility at the logic level. (c) Subtask Planner: performs motion planning for the subtasks in the actual plan to ensure the feasibility of the plan at the motion level. (d) Robot Controller: generates control sequences based on the first task in the plan and the current state. (e) State Evaluator: Evaluates numerical parameters based on sensor information and applies predicate resolution rules to evaluate logic state.}
\label{framework}
\end{figure*}

Our core idea is that the validity of a plan presupposes its validity at both the logic and motion levels. Therefore, to ensure the feasibility of the plan, real-time replanning is essential at both the logic and motion levels during execution. We construct an modular multi-Level replanning framework, as shown in Fig. \ref{framework}, which consists of five components. The \textbf{TAMP Solver} plans a nominal subtask plan based on the logic and numerical states. The \textbf{Subtask Scheduler} reconstructs a actual plan based on the nominal plan in real time. The \textbf{Subtask Planner} iteratively plans motion paths for each subtask based on the latest state. The \textbf{Robot Controller} executes the first subtask in the actual plan based on the planning results of the Subtask Planner and the latest state. The \textbf{State Evaluator} refines the numerical state of each object based on the sensor information and parses the logic state according to the predicate parsing rules.

The workflow and principles of these five components are described in detail below.
\subsection{TAMP Solver}
The function of the TAMP Solver is to generate a nominal subtask plan ${P_{n}} = ({\tau _1},{\tau _2},{\tau _3}, \cdots ,{\tau_n})$ based on the domain description file, user goals and current state.

We choose PDDLStream\cite{garrett2020pddlstream} as our TAMP Solver, which can generate plans with \texttt{action} as elements. Each \texttt{action} contains \texttt{parameters}, \texttt{precondition} and \texttt{effect} that correspond to a segment of a motion path. To enable the framework to perform low-level replanning quickly, we propose the concept of \texttt{subtask} $\tau (type,params,{L_P},{L_E},f_{target},{\pi_n},u_n,\chi_{end})$ based on \texttt{action}. As an example, the \texttt{place} subtask in Fig. \ref{interference}\subref{heavy}  has the following form.\\

\noindent\{\textbf{type}: \texttt{place}

\noindent\textbf{params}: \texttt{($b_G, b_R, {}^R{T_G}$)}

\noindent\bm{${L_P}$}: \texttt{\{(Holding $b_G$),(GripperOn $b_R$)\}}

\noindent\bm{${L_E}$}: \texttt{\{(On $b_G$ $b_R$)\}}

\noindent\bm{$f_{target}$}: \texttt{$f_{target}(\chi_{start},params)$}

\noindent\bm{$\pi_n$}: \texttt{$\pi_n(\chi_{start},\chi_{target})$}

\noindent\bm{$u_n$}: \texttt{($q(t),\dot q(t),\ddot q(t)$)}

\noindent\bm{$\pi_a$}: \texttt{$\pi_a(\chi_{curr},u_n)$}

\noindent\bm{$\chi_{end}$}: \texttt{\{$b_G$:((0.3,0,0.075),(0,0,0,1)),$b_R$:$ \cdots$\}}\}\\

\bm{$type$} describes the operation of the subtask. \bm{$params$} contains the manipulation object and numerical parameters, primarily object-centric parameters. For example, ${}^G{T_B}$ denotes the placement of the blue block relative to the green block, enabling the robot to calculate the corresponding placement after moving the green block. \bm{${L_P}$} and \bm{${L_E}$} are the logic state sets of preconditions and effects, which will be used for logic replanning in the Subtask Scheduler. \bm{$f_{target}$} will generate target state ${\chi_{target}}$ of this subtask based on the start state ${\chi_{start}}$ and $params$, \bm{$\pi_n$} is the motion planning algorithm that will generate a nominal motion path \bm{$u_n$} and end state \bm{$\chi_{end}$} according to $\chi_{start}$ and $\chi_{target}$. These processes will be further explained in the Subtask Planner.

\subsection{Subtask Scheduler}
Ideally, after the execution of the previous subtasks, the preconditions of the next task will be satisfied. However, in the presence of interference, the logic state may change, leading to the subsequent subtasks not being executed. In such cases, the Subtask Scheduler can instruct the robot to backtrack to the previous subtasks, skip some subtasks, or notify the TAMP Solver for replanning. The Subtask Scheduler rearranges subtasks from the nominal plan $P_n$ in real-time to construct an actual plan ${P_{a}} = ({t_1},{t_2},{t_3}, \cdots ,{t_m})$ that can reach the goal logic state $L_{goal}$ from the current logic state $L_{curr}$. The actual plan $P_{a}$ needs to satisfy the following constraints.

\begin{align}
 L_P^{i + 1}\subseteq \sum\limits_{k = 0}^i {L_E^k } ,  \forall i \in [0,m], 
{t_i} \in \{ {\tau _1},{\tau _2},{\tau _3}, \cdots ,{\tau _n}\}
\end{align}

The superscript of $L$ represents that the parameter belongs to the subtask $t_i$. $L_{E}^{0}$ stands for the current logic state $L_{curr}$, and $L_{P}^{m+1}$ stands for the goal logic state $L_{goal}$. To ensure the feasibility of $P_a$  at the logic level, the preconditions of the $i+1$ th subtask $L_{P}^{i+1}$ need to be valid after the execution of the first $i$ subtasks, and the goal state $L_{goal}$ needs to be satisfied after the execution of the whole plan $P_a$. Ideally, the actual plan $P_a$ is equal to the nominal plan $P_n$ initially and becomes shorter as the subtasks continue to be completed.

We propose an algorithm to efficiently refine an actual plan from the nominal plan, and its pseudo-code is shown in Alg. \ref{alg:subtask_scheduler}. The Subtask Scheduler cyclically traverses the nominal plan in reverse order, finds the first subtask whose precondition is satisfied, adds it to the actual plan $P_a$, and then removes it from the nominal plan $P_n$. The Subtask Scheduler modifies the cumulative logic state $L_{cumu}$ according to its effect and returns if $L_{cumu}$ reaches the goal state $L_{goal}$; otherwise it restarts the traversal. There may be multiple subtasks that can be executed in a traversal, but only the last one is added because it is closest to the goal. This is why the algorithm traverses in reverse order.

\begin{algorithm}[!ht]
    \begin{algorithmic}[1]
        \Require{$L_{curr}$, $L_{goal}$, $P_{n}$}
        \Ensure{$P_{a}$}

        \Procedure{Refine}{$L_{curr}$, $L_{goal}$, $P_{n}$}
            \State $P_{temp}\leftarrow reverse(P_{n})$ 
            // $({\tau _n},{\tau _{n - 1}}, \cdots ,{\tau _1})$
            \State $L_{cumu}\leftarrow L_{curr}$
            \State $P_{a}\leftarrow \emptyset$ 
            \While{not $L_{goal}\subseteq L_{cumu}$}
                \State $flag = TRUE$       
                \For{$\tau \in P_{temp}$}
                    \If{$\tau.L_{P} \subseteq L_{cumu} $}
                        \State $P_{temp}.pop(\tau)$
                        \State $P_{a}.push(\tau)$
                        \State $L_{cumu} = L_{cumu} \cup  \tau.L_{E}$
                        \State $flag = FALSE$
                        \State \textbf{break} 
                    \EndIf                    
                \EndFor                
                \If{$flag$}
                    \State \textbf{return} None
                \EndIf                              
            \EndWhile
            \State \textbf{return} $P_{a}$
        \EndProcedure
    \end{algorithmic}
    \caption{Refine an actual subtask plan}
    \label{alg:subtask_scheduler}
\end{algorithm}

If no feasible subtasks are found in a traversal, or if all subtasks are selected but still do not reach the goal, it indicates that the nominal plan $P_n$ has failed at the logic level. In such cases, the Subtask Scheduler will notify the TAMP Solver to start replanning. If a feasible actual plan is found, the Subtask Scheduler will send the plan $P_a$ to the Subtask Planner to verify the motion feasibility of the plan.

\subsection{Subtask Planner}
The Subtask Planner iterates over the actual plan $P_a$ and sequentially plans the motion of the subtasks. The Subtask Planner not only attempts to find a feasible motion online to verify whether the actual plan $P_a$ is valid at the motion level but also continuously optimizes the result.

The planning process of each subtask is divided into two steps. First, the target state $\chi_{target}$ is obtained by $f_{target}$ based on the start state $\chi_{start}$ and $params$. Second, according to the planning algorithm $\pi_n$, the nominal motion path $u_n$ from the start state $\chi_{start}$ to reach the target state $\chi_{target}$ ,the environment state $\chi_{end}$ at the end moment of the planning will be recorded.

\begin{figure}[htbp]
\centering
\subfloat[]{\includegraphics[width=0.96\columnwidth]{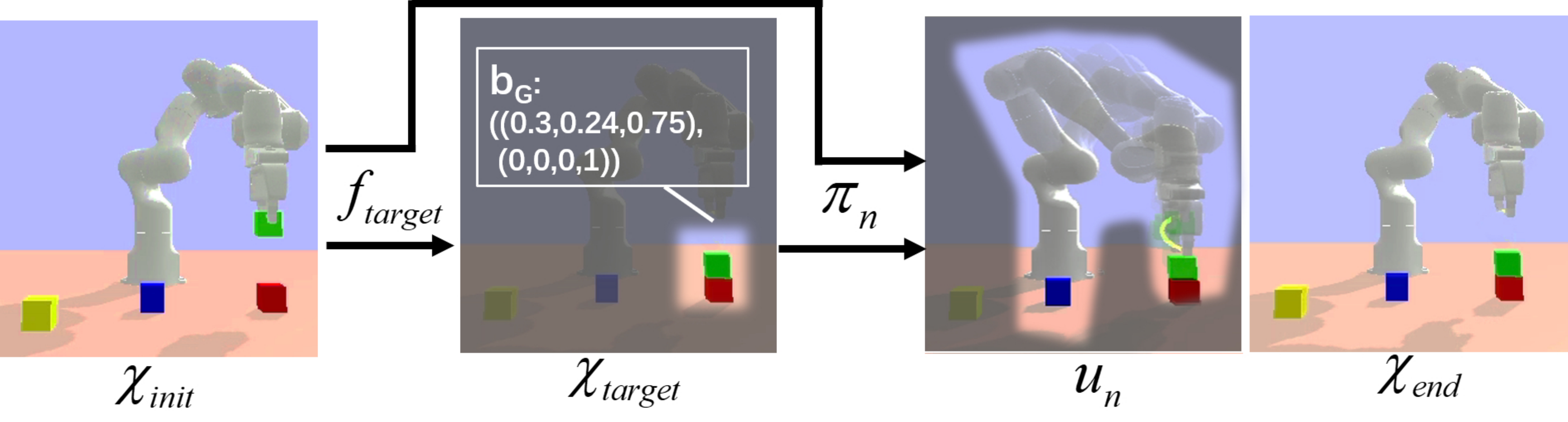}
\label{place}
}%

\subfloat[]{\includegraphics[width=0.96\columnwidth]{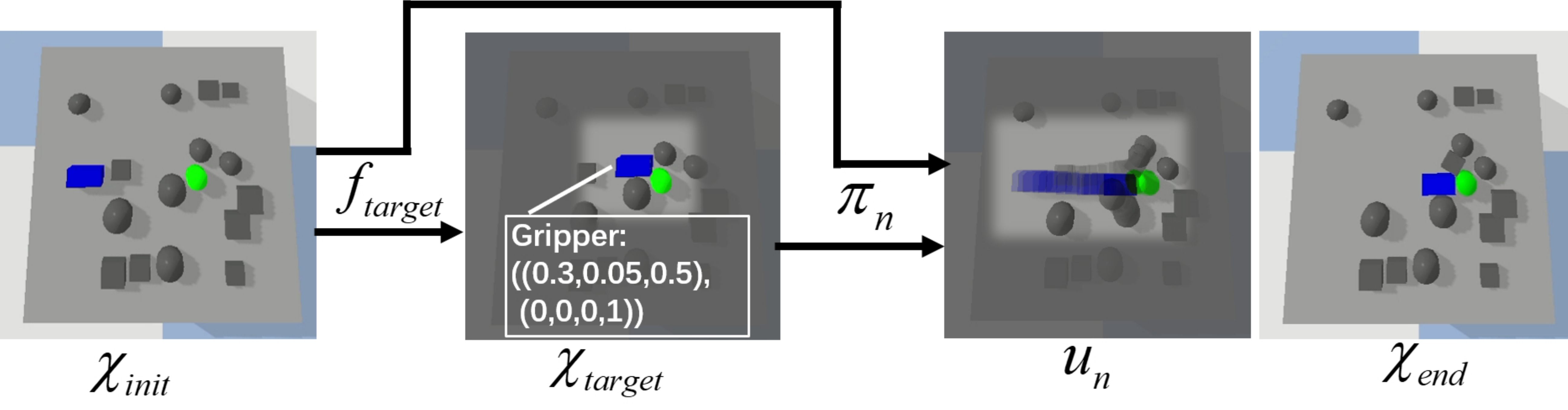}
\label{push}
}%

\caption{The workflow of the Subtask Planner. First generate the target state and then plan the robot motion path and obtain the end state. (a) \texttt{place} subtask. (b) \texttt{push} subtasks}
\label{subtask_planner}
\end{figure}

Take \texttt{place} as an example, as shown in Fig. \ref{subtask_planner}\subref{place}, the target of this subtask is to place green block on top of red block. So $f_{target}$ represents the relative placement position ${}^R{T_G}$ sample function in \texttt{place} subtask. The target state $\chi_{target}$ is the absolute position of the green block calculated based on ${}^R{T_G}$. 

Subtasks are replanned multiple times during execution. Before planning, Subtask Planner will verify ${}^R{T_G}$ according to the start state $\chi_{start}$. If target position of the green block causes a collision, the Subtask Planner will resample a new relative placement pose ${}^R{T_G}$, otherwise, the previous parameters will be retained.

The motion planning algorithm $\pi_n$ of \texttt{place} is divided into two steps. Firsty, the inverse kinematics is used to obtain the robot placement configuration, and then a collision-free joint angle sequence $u_n$ is planned from the current configuration to the placement configuration. In the planning process, if the previous result $u_n$ remains feasible, planner Subtask Planner full re-planning and instead attempts to smooth and optimize the existing path.

In execution, the robot returns along the original path after placing the block, so only the position of the green block changes in the end state $\chi_{end}$ compared to the start state $\chi_{start}$.

For other subtasks, not only the state of the target object may be changed, but also the states of some unrelated objects. As shown in Fig. \ref{subtask_planner}\subref{push}, robot need to push its gripper to the target. First $f_{target}$ samples a feasible relative position. Then, $\pi_n$ plans the velocity trajectory of the joints so that the gripper can reach the target. When the subtask is completed, not only the gripper, but also the target and unrelated objects have been changed.

\begin{algorithm}[!ht]
    \begin{algorithmic}[1]
        \Procedure{Plan}{}
            \While{\textbf{True}}
                \For{$\tau \in P_{a}$}                    
                    \State $\chi_{start} = \tau_{last}.\chi_{end}$
                    \State $\chi_{target} =\tau.f_{target}(\tau.params,\chi_{start})$
                    \If{$\chi_{target}$ is None}
                        \State \textbf{return} False
                    \EndIf
                    \State $\tau.\chi_{end},\tau.u_n =\tau.\pi_n(\chi_{target},\chi_{start},\tau.u_n)$
                    \If{$\tau.u$ is None}
                        \State \textbf{return} False
                    \EndIf
                \EndFor
            \EndWhile            
        \EndProcedure
    \end{algorithmic}
    \caption{Subtask Planner Workflow}
    \label{alg:subtask_planner}
\end{algorithm}

The workflow of the Subtask Planner is outlined in Alg. \ref{alg:subtask_planner}. To ensure continuity at the motion level, the start state $\chi_{start}$ of each subtask $\tau$ is set as the end state $\chi_{end}$ of its previous subtask $\tau_{last}$. For the first subtask, the start state is the current environment state $\chi_{curr}$. During motion planning, the previous result $u_n$ can serve as the initial solution, accelerating the planning process and continuously optimizing the result in the absence of interference.

If any subtask experiences motion planning failure, it implies that the actual plan $P_a$ is infeasible at the motion level. The Subtask Planner then notifies the TAMP Solver for replanning. If all subtasks pass planning successfully, the Subtask Planner sends the planning result of the first subtask to the Robot Controller for execution.

For some deterministic subtasks, such as \texttt{place}, the end state $\chi_{end}$ can be obtained based on $f_{target}$. Then, motion planning of the next subtask can commence, and the framework can utilize the multi-core capabilities of the CPU to plan multiple subtasks in parallel thereby increasing the planning frequency. 

\subsection{Robot Controller}
The Robot Controller will generate the actual inputs $u_a$ of the robot based on the planning result of the first subtask $u_n$ and current state $\chi_{curr}$. The Robot Controller can use different control strategies, e.g., for \texttt{pick} and \texttt{place} subtask, the Robot Controller can use the Cartesian impedance method to reduce unexpected contact forces. For \texttt{move} subtask, Robot Controller can use RMP\cite{cheng2021rmpflow} for reactive control.

\begin{align}
u_{a} = \pi_a (\chi_{curr},{u_{n}})
\end{align}

\begin{figure}[htbp]
\centering
\includegraphics[width=1.0\columnwidth]{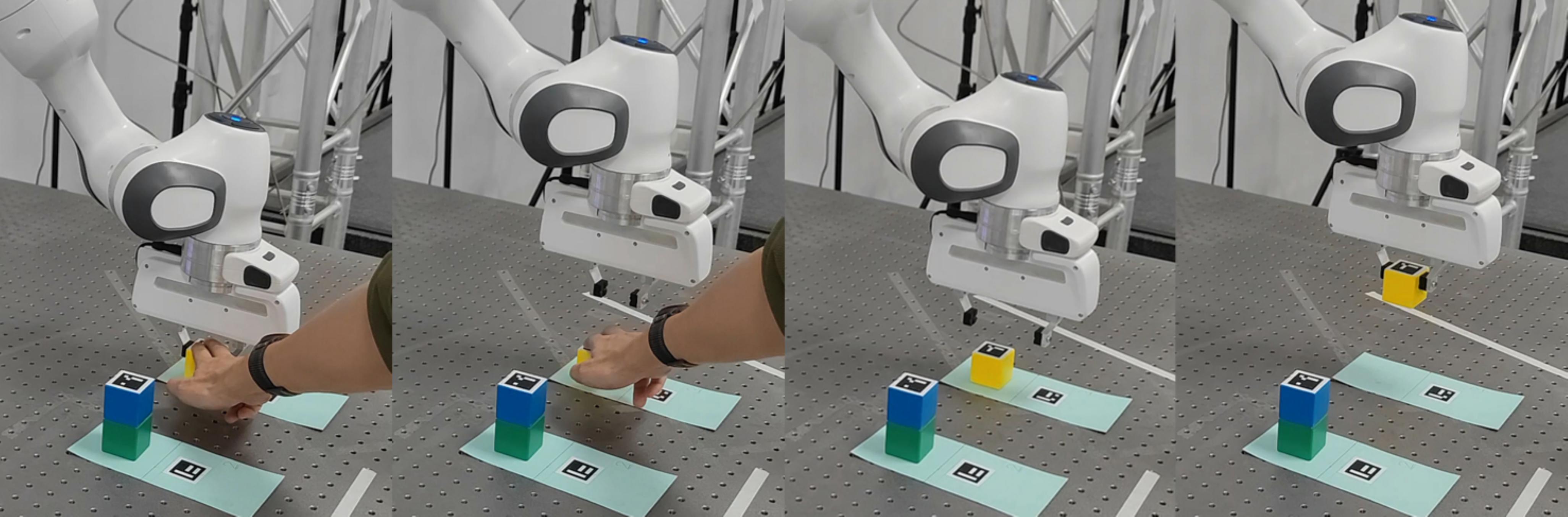}
\caption{The execution of the \texttt{pick} subtask failed due to the sudden obstruction of the target block. However, this subtask is still feasible at the motion level. After repeated execution, the yellow block was successfully grabbed.}
\label{execution_failure}
\end{figure}

The robot adapts to the dynamic environment by executing the first subtask in the actual plan $P_a$. Execution Failure prompts the robot to exit the current subtask, and such a failure does not trigger TAMP replanning, as it does not represent a motion level failure of the entire plan. For example, as shown in Fig. \ref{execution_failure}, in the \texttt{pick} subtask, at the moment when the arm attempts to grab, the experimenter blocks the red block with his hand, causing the execution of this subtask to fail. However, this subtask remains feasible at the motion level, and the Robot Controller will attempt to execute this subtask again in the next execution.

Furthermore, the first subtask in the actual plan $P_a$ may change while the robot is in operation. To prevent oscillations caused by abrupt switches between subtasks, the robot does not switch the current subtask once execution has commenced.

\subsection{State Evaluator}
\begin{figure}[htbp]
\centering
\includegraphics[width=1.0\columnwidth]{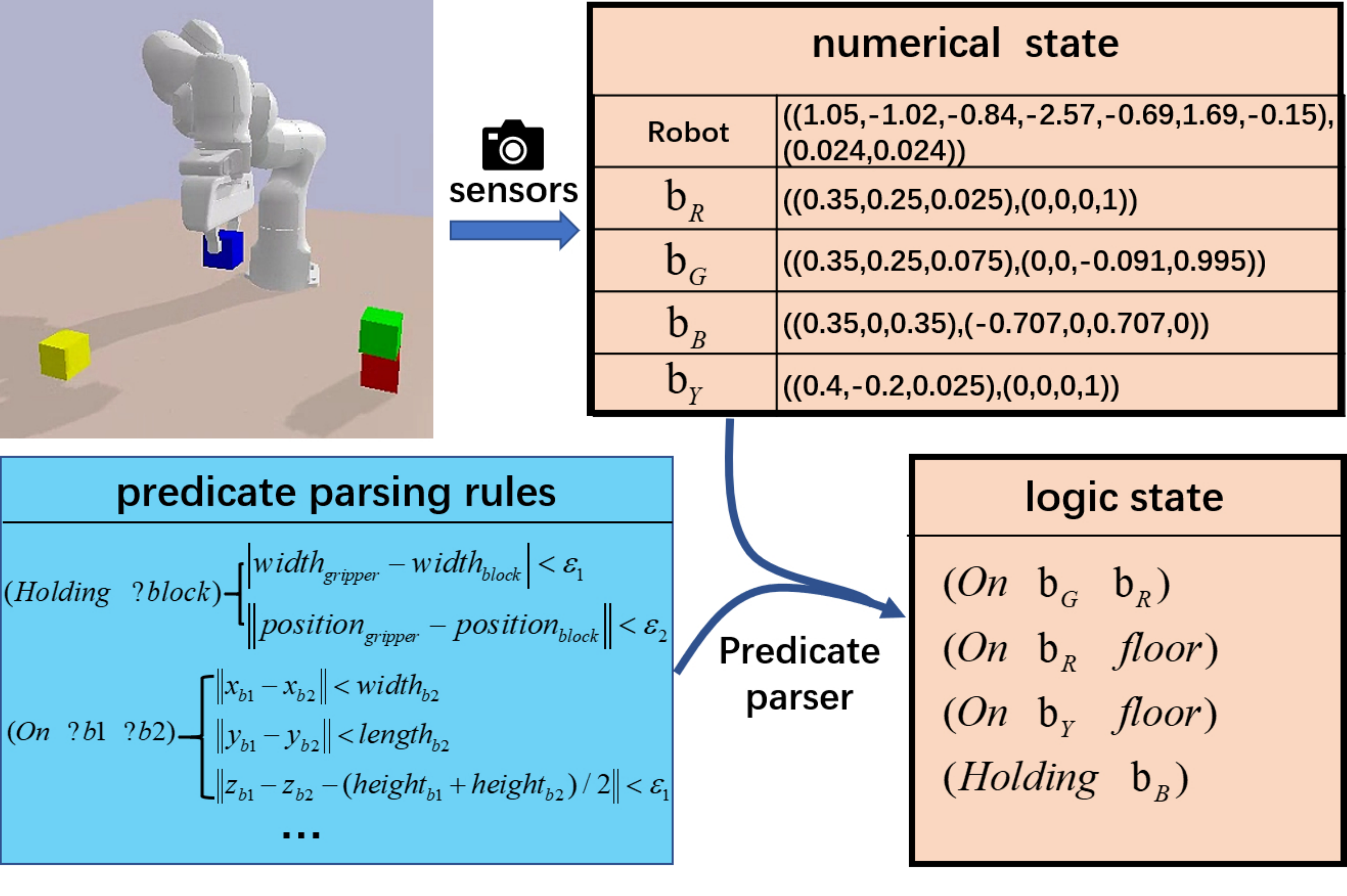}
\caption{The workflow of the State Evaluator. The State Evaluator first extracts the numerical state based on the sensors and then applies predicate parsing rules to parse out the logic state.}
\label{state_evaluator}
\end{figure}
The workflow of the State Evaluator is divided into two phases, as shown in Fig. \ref{state_evaluator}. In the first phase, the State Evaluator extracts the numerical state of the current environment using the information from various sensors. This involves reading parameters from encoders, torque sensors, or estimating object positions based on RGBD images.
In the second stage, the State Evaluator refines the logic state based on the numerical state and predicate parsing rules. The logic state consists of a series of predicates, describing the semantic state of objects and are mathematically represented as a set of inequalities. For example, the predicate \texttt{(Holding ?block)}, which represents whether the robot is holding a certain block, corresponds to the following set of inequalities:
\begin{align}
 \mid  width_{\text {gripper }}-  width_{\text {block}} \mid<\varepsilon_{1} \\
 \|  p_{\text{gripper }}-  p_{\text{block}} \|<\varepsilon_{2}  
\end{align}
During parsing, the State Evaluator traverses all possible object combinations for each predicate and adds the predicates that satisfy the set of inequalities to the current logic state $L_{curr}$.
For \texttt{(Holding ?block)}, the State Evaluator iterates through the numerical states of all \texttt{(Gripper, block)} combinations. If a combination such as \texttt{(Gripper, block$_{i}$)} satisfies the aforementioned set of inequalities, then \texttt{(Holding block$_{i}$)} is appended to the current logic state $L_{curr}$.

\section{EXPERIMENT AND RESULTS}
\subsection{Experiment Domain}
The experiment contains two scenarios, the rearrange and stack domains. In the rearrange domain, the robot needs to place four of the five blocks into each of the two regions. In the stack domain, the robot needs to stack three of the four blocks in a specific order, which is more challenging. This is because in rearrange domain the blocks are usually independent of each other, while in stack domain, the blocks are placed in a strict order and are more likely to be disturbed in the logic level.
These two domains share a common set of primitives and contain the following four types of subtasks.

\begin{itemize}
    \item \texttt{move\_free(b)}
    \item \texttt{pick(b)}
    \item \texttt{move\_hold(b$_1$, b$_2$, ${}^2{T_1}$)}
    \item \texttt{place(b$_1$, b$_2$, ${}^2{T_1}$)}
\end{itemize}

The target of \texttt{move\_free} and \texttt{move\_hold} is to move the gripper over the target object; the difference is that the \texttt{move\_free} presupposes an empty gripper, whereas \texttt{move\_hold} clamps object. \texttt{pick} is to pick up the object, and \texttt{place} is to place the object in a specific relative position to the target.

We have also designed three levels of interference: 

\begin{itemize}
\item \textbf{light interference}: Push the block away from its original position before grabbing, causing the \texttt{pick} subtask to fail.

\item \textbf{Mid interference}: After placing the block, when the robot is ready to grab the next block, the block is placed back in its original position.

\item \textbf{heavy interference}: Place unrelated objects on a target block at the beginning of the task.

\end{itemize}

\begin{figure*}[htbp]
\centering
\includegraphics[width=2.0\columnwidth]{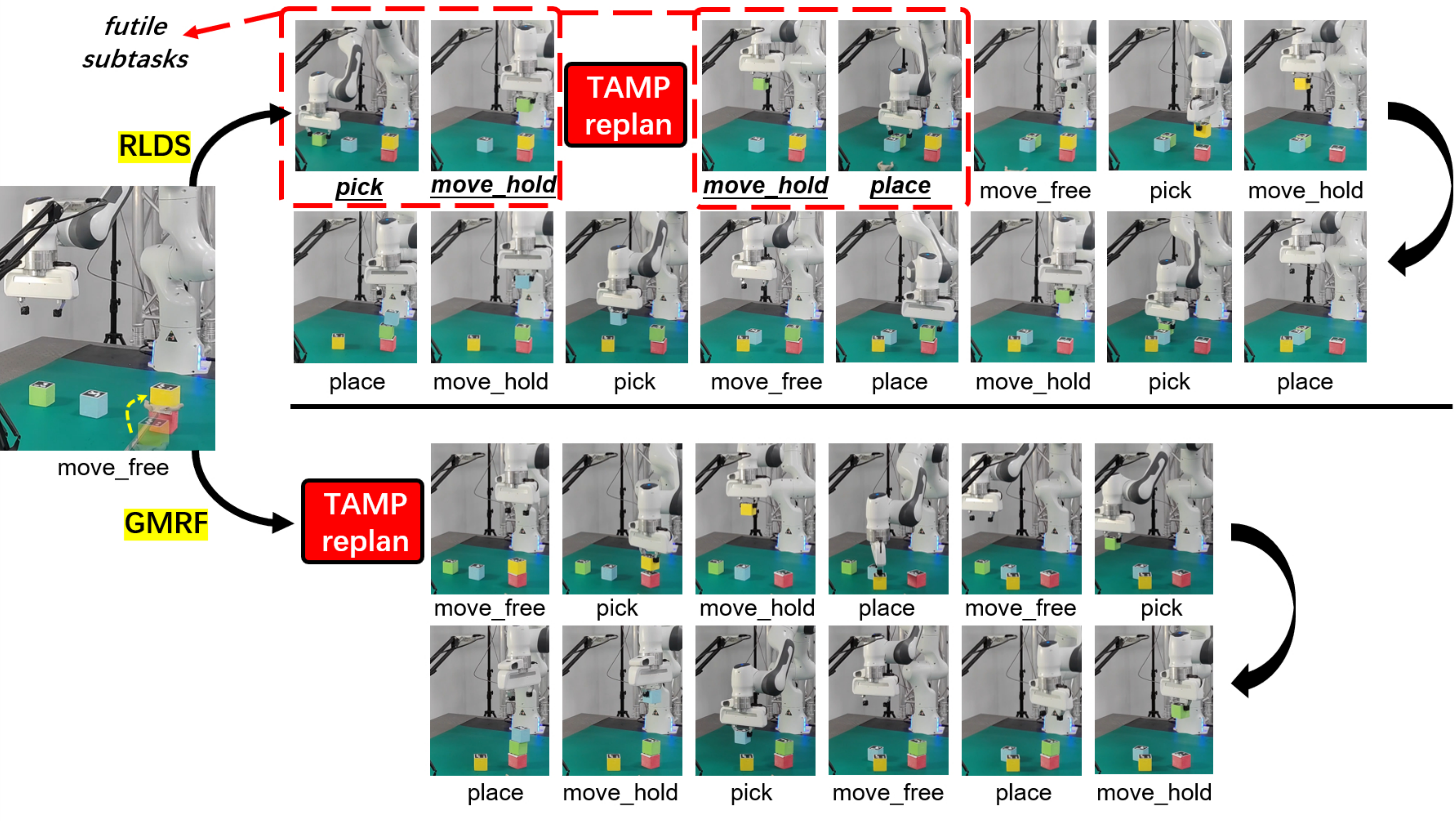}
\caption{Screenshots of the execution sequence of the RLDS and MMRF performing the stack task under heavy interference. After placing the yellow block , the Subtask Planner in MMRF verified that the current plan was infeasible at the motion level and triggered TAMP replanning. In contrast, RLDS determined that the plan was infeasible only when attempting to place the green block. This led to RLDS executing four more subtasks than MMRF.}
\label{RLDS_and_MMRF}
\end{figure*}
\subsection{Experiment Setting}
We choose PDDLStream\cite{garrett2020pddlstream} as the TAMP Solver. Since it is based on Pybullet\cite{coumans2016pybullet}, the framework synchronizes the simulation environment according to the state of the real world. The motion planning algorithm for these four types of subtasks in the Subtask Planner is RRTconnect\cite{kuffner2000rrt}. The Robot Controller performs fifth-degree polynomial interpolation on the nominal path based on the robot state, and utilizes joint impedance control to track the trajectory.

We select Franka Emika Panda for these experiments. RGB images are collected using a D435i camera located outside the robot, and the positions of blocks are estimated by ArUco markers on them. ROS is utilized for the communication of the framework with the Panda and the D435i. The framework runs on a laptop with \textit{Intel Core i5 10210U 1.6GHz} CPU and 16GB of RAM.

To avoid failures caused by sensor noises, the TAMP Solver only replans if the Subtask Scheduler does not find a feasible plan within twenty seconds or if the Subtask Planner does not find a motion path of each subtask within ten seconds.

\subsection{Real-world Experiment}
Three frameworks have been designed to reflect the importance of real-time planning at the level of operational logic and motion.

\textbf{Reactive Control(RC)}: RC contains only the Robot Controller and the State Evaluator. The Robot Controller linearly executes the nominal plan. It starts executing the next subtask when its preconditions are satisfied. Otherwise it repeats the planning and execution of the current subtask.

\textbf{Robust Logical Dynamic System(RLDS)}: The original RLDS\cite{paxton2019representing} is approximately a combination of Task Scheduler, Robot Controller, and State Evaluator. Therefore, the task plan of RLDS is defined manually and cannot be generated autonomously. In order to demonstrate the importance of the Subtask Planner, a TAMP solver is added to RLDS.

\textbf{Modular Multi-Level Replanning Framework(MMRF)}: the framework proposed in this paper, allows online replanning at both logic and motion levels.

These three frameworks are executed five times each in two task domains and three interference conditions. Each task domain has the same start and end conditions, and the initial layout of the objects is essentially the same. The success rate and completion time of each framework under different interference were then recorded. The completion time does not include the initial TAMP planning time.

\subsection{Result Analysis}
Table \ref{result} shows results from experiments. RC can cope with slight interference with the help of repeated execution. However, RC can only determine whether it can proceed to the next subtask or not. This limitation results in the inability of RC to complete the task under middle interference. In the rearrange domain, where the operations on different blocks are logically independent, the robot can execute the plan sequentially, but it does not replace the moved objects. In the stack domain, the placement of the top block is invalidated when the second block is moved out of the way. As a result, RC has a 0\% completion rate with the middle and heavy interference.

\begin{table}[htbp]
    \caption{Completion Times of  experiments}
    \centering
    \begin{tabular}{llccc}
        \toprule
        \multirow{2}*{Domain}& \multirow{2}*{Framework}& \multicolumn{3}{c}{Completion Time(s)}  \\
        &       &\multicolumn{1}{c}{slight}&\multicolumn{1}{c}{middle}&\multicolumn{1}{c}{heavy}\\
        \midrule
        \multirow{3}*{rearrange}& RC& 112.1 $\pm$ 12.3&n/a&  n/a\\
        & RDLS& 115.4 $\pm$ 12.6& 119.3 $\pm$ 15.6& 159.5 $\pm$ 21.8\\
        & MMRF& 103.1 $\pm$ 9.6& 105.5 $\pm$ 10.8& 136.8 $\pm$ 15.3\\
        \midrule
        \multirow{3}*{stack}& RC& 69.3 $\pm$ 9.8& n/a& n/a\\
        & RDLS& 65.4 $\pm$ 9.1& 81.2 $\pm$ 11.8& 138.1 $\pm$ 17.8\\
        & MMRF& 57.4 $\pm$ 7.3& 70.7 $\pm$ 8.4& 95.2 $\pm$ 12.5\\
        \bottomrule
    \end{tabular}   
    \label{result}
\end{table}

RDLS can adjust the execution order to complete the task quickly under middle interference and can regenerate the plan using the TAMP Solver under heavy interference,  resulting in a 100\% success rate across all scenarios. However, due to the absence of the Subtask Planner, RLDS needs to replan the motion path at the beginning of each subtask, leading to short pauses between subtasks. As a result, the completion times of RLDS are all longer than those of MMRF.
It is worth noting that under heavy interference, the difference between the two completion times is significantly larger in the stack domain than in the rearrange domain.

This is due to the absence of a Subtask Planner, which prevents RLDS from detecting that some subtasks cannot be completed at the motion level until RLDS is about to execute them. This delay in detection has little impact in rearrange because the subtasks are relatively independent. However, in stack domain, this can make the manipulation after heavy interference all for naught. As shown in Fig. \ref{RLDS_and_MMRF}, we place a yellow block on red block during the first planning process of the TAMP Solver, making the subtask of placing green block infeasible at the motion level. The RLDS still picks the green block up and does not detect the subtasks as infeasible until it is about to place. In contrast, the Subtask Planner in MMRF parallelly plans the motion paths of all subtasks, detects that the subtask with green blocks placed is infeasible at the motion level, then notifies TAMP Solver to conduct a replanning to generate a new plan. This delay results in RLDS executing four more subtasks than MMRF.

\begin{figure}[htbp]
\centering
\subfloat[]{\includegraphics[width=0.45\columnwidth]{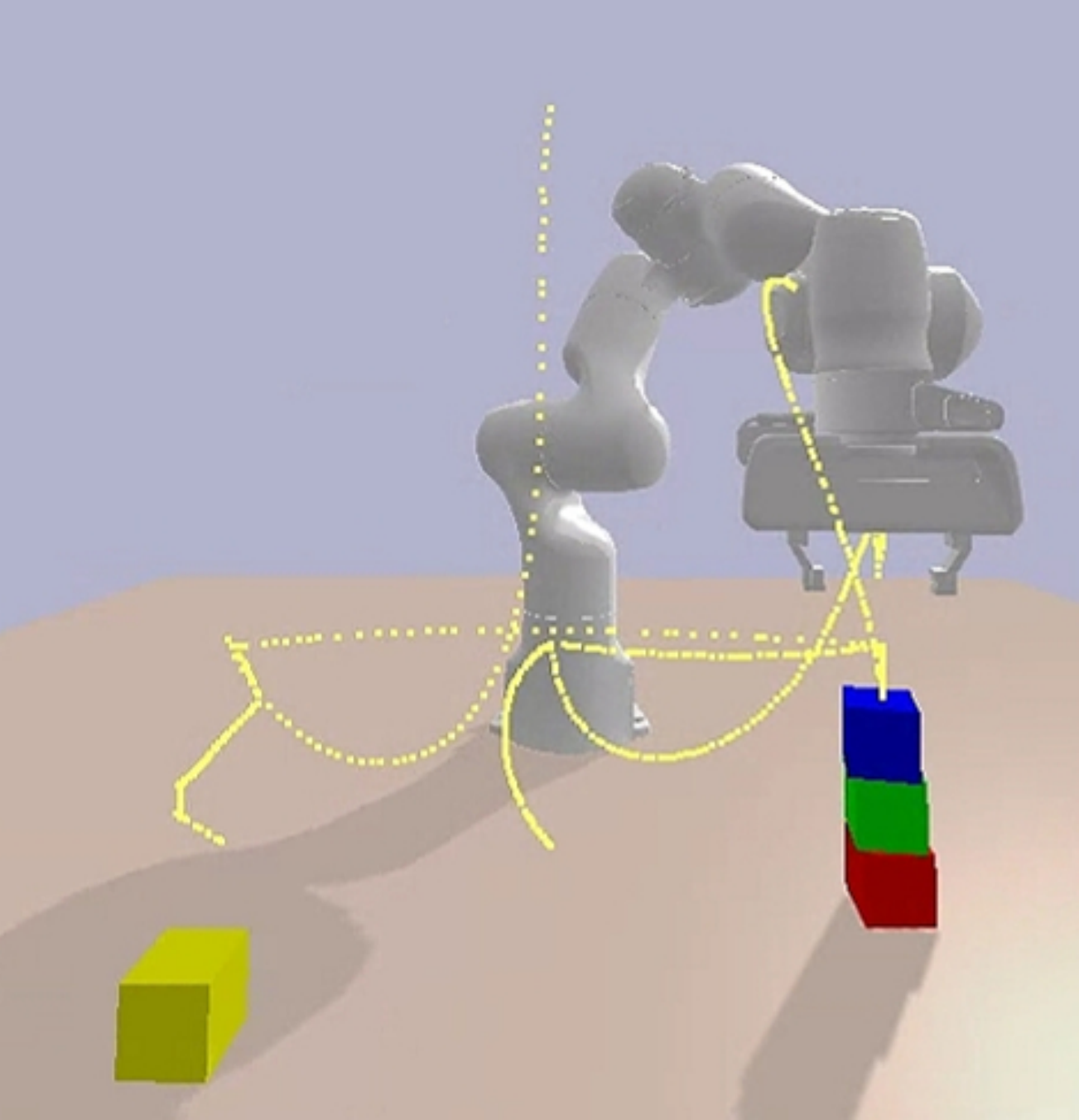}
\label{RLDS}
}
\subfloat[]{\includegraphics[width=0.45\columnwidth]{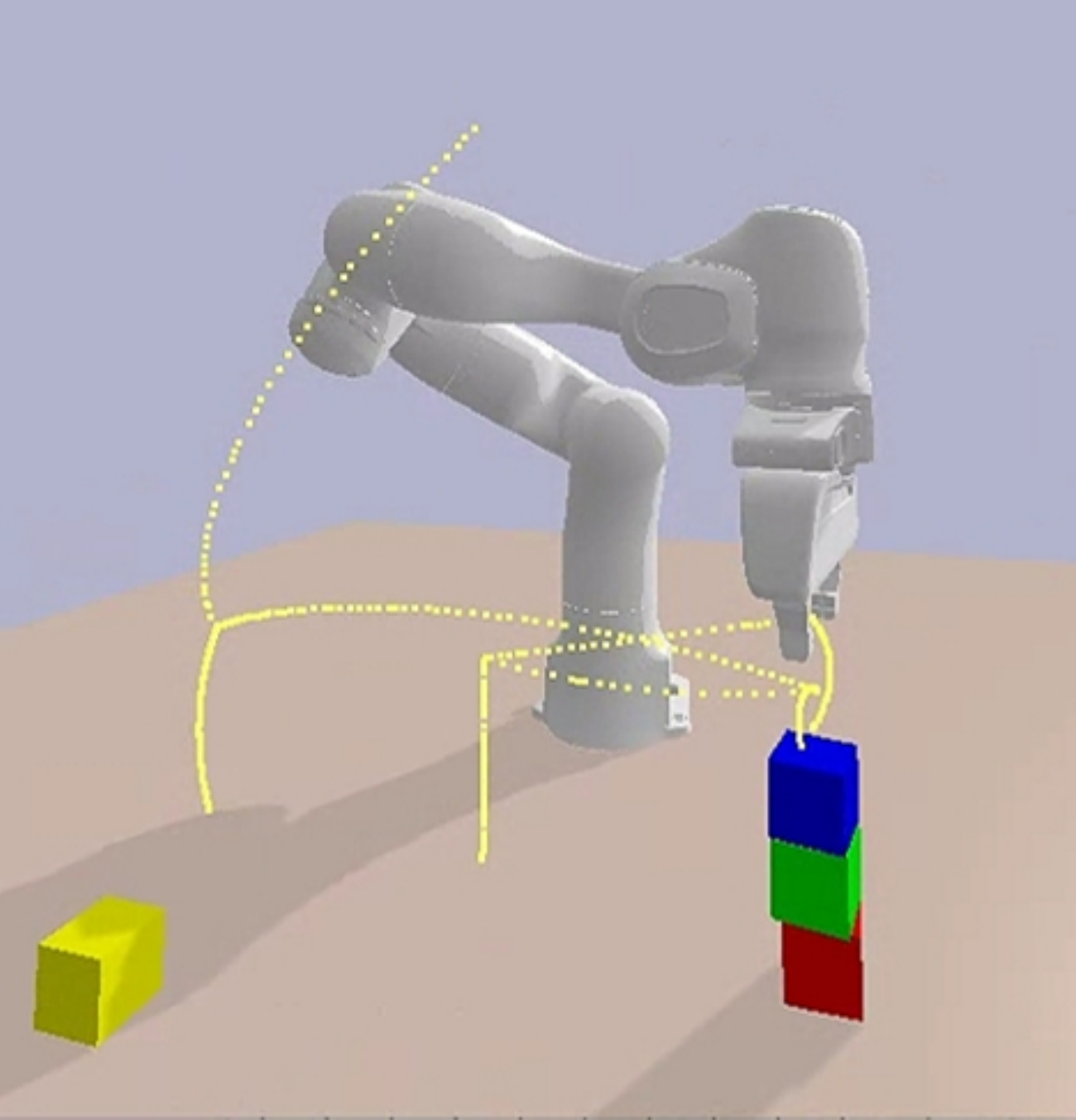}
\label{MMRF}
}
\caption{Trajectories of (a) RLDS and (b) MMRF in a stack task without interference. The trajectory of MMRF is better than that of RLDS under online optimization by the Subtask Planner.}
\label{optimization}
\end{figure}

MMRF has a 100\% completion rate for all three interference and has shorter completion times than the other groups for all scenarios. 
Two reasons contribute to this success. First, the online motion planning of the Subtask Planner increases the likelihood that the initial path obtained by the Robot Controller is a feasible path, thereby significantly reducing the planning time for the Robot Controller. Second, after finding the feasible paths, the Subtask Planner continuously optimizes the paths, which is similar to AnyTimeRRT\cite{karaman2011anytime} in terms of effect.

As shown in Fig. \ref{optimization}, the yellow dashed lines represent the end motion trajectories of the robot in the stack domain. Fig. \ref{optimization}\subref{RLDS} shows the trajectory of RLDS, resembling a typical path planned by RRT—long and winding. And Fig. \ref{optimization}\subref{MMRF} shows the trajectory of MMRF, markedly improved compared to RLDS after online optimization.

\section{CONCLUSION AND FUTURE WORK}

In this paper, we first summarize three kinds of interference that the TAMP algorithm suffers from in practical applications. In order to cope with these interference quickly, we propose a modular multi-level replanning TAMP framework. The framework generates a nominal plan by the TAMP Solver. During execution, it conducts real-time replanning at both the logic and motion levels based on the current state. The low-level online replanning significantly reduces the number of time-consuming TAMP replanning. 

We experimentally demonstrate that our framework reduces completion time by an average of 13\% compared to the traditional replanning framework under slight and middle interference due to the online optimization of the Subtask Planner. In scenarios with heavy interference, the reduction in completion time reaches 28\%. This advantage becomes more pronounced with longer task sequences and less independent subtasks.

The future directions of work are as follows:

Use visual-language-model(VLM) as a way to parse logic states. First, handwritten rules are very cumbersome and difficult to cover all states; second, the computation time to refine logic states now grows exponentially as the number of objects and predicates increases.

Optimize the scheduling algorithm to generate the shortest task path. Alg. \ref{alg:subtask_scheduler} is only a feasible method, but it does not guarantee the generation of the shortest plan. The algorithm should be improved to pursue the fastest completion of tasks under interference.

\bibliographystyle{IEEEtran}
\bibliography{ref}
\end{document}